\documentclass[letterpaper]{article} 
\usepackage[]{aaai24}  
\usepackage{times}  
\usepackage{helvet}  
\usepackage{courier}  
\usepackage[hyphens]{url}  
\usepackage{graphicx} 
\usepackage{natbib}  
\usepackage{caption} 
\usepackage{amssymb}
\usepackage{xcolor}
\definecolor{citecolor}{HTML}{2980b9}
\usepackage{multirow}
\usepackage[utf8]{inputenc} 
\usepackage[T1]{fontenc}    
\usepackage{url}            
\usepackage{booktabs}       
\usepackage{amsfonts}       
\usepackage{nicefrac}       
\usepackage{microtype}      
\usepackage{amsmath}
\usepackage{graphicx} 
\usepackage{adjustbox}
\usepackage{xcolor}         
\usepackage{color, colortbl}

\makeatletter
\newcommand\figcaption{\def\@captype{figure}\caption}
  \newcommand\tabcaption{\def\@captype{table}\caption}
\makeatother

\usepackage{subcaption}
\usepackage{makecell} 
\usepackage[capitalize]{cleveref}
\crefname{section}{Sec.}{Secs.}
\Crefname{section}{Section}{Sections}
\Crefname{table}{Table}{Tables}
\crefname{table}{Tab.}{Tabs.}
\usepackage{algorithm}
\usepackage{algorithmic}

\usepackage{newfloat}
\usepackage{listings}
\DeclareCaptionStyle{ruled}{labelfont=normalfont,labelsep=colon,strut=off} 
\floatstyle{ruled}
\newfloat{listing}{tb}{lst}{}
\floatname{listing}{Listing}
\title{Referred by Multi-Modality:\\ A Unified Temporal Transformer for Video Object Segmentation}
\author{
    Shilin Yan\textsuperscript{1,2}\equalcontrib, Renrui Zhang\textsuperscript{2,3}\equalcontrib, Ziyu Guo\textsuperscript{3}\equalcontrib, Wenchao Chen\textsuperscript{1}, Wei Zhang\textsuperscript{1$\dagger$}, Hongyang Li\textsuperscript{2$\dagger$}\\Yu Qiao\textsuperscript{2}, Hao Dong\textsuperscript{4,5}, Zhongjiang He\textsuperscript{6}, Peng Gao\textsuperscript{2}\thanks{Corresponding Author.}\vspace{0.1cm}\\
}
\affiliations{
    \textsuperscript{1}School of Computer Science, Fudan University,\quad 
    \textsuperscript{2}Shanghai Artificial Intelligence Laboratory,\\
    \textsuperscript{3}The Chinese University of Hong Kong,\quad
    \textsuperscript{4}School of CS, Peking University,\quad \textsuperscript{5}PKU-agibot Lab\\
    \textsuperscript{6}China Telecom Corporation Ltd. Data\&AI Technology Company\vspace{0.1cm}\\
    {tattoo.ysl}@gmail.com, \{zhangrenrui, guoziyu, gaopeng\}@pjlab.org.cn
}

\usepackage{bibentry}

\begin{document}

\maketitle

\begin{abstract}

Recently, video object segmentation (VOS) referred by multi-modal signals, e.g., language and audio, has evoked increasing attention in both industry and academia. It is challenging for exploring the semantic alignment within modalities and the visual correspondence across frames.
However, existing methods adopt separate network architectures for different modalities, and neglect the inter-frame temporal interaction with references. In this paper, we propose \textbf{MUTR}, a \textbf{M}ulti-modal \textbf{U}nified \textbf{T}emporal transformer for \textbf{R}eferring video object segmentation. With a unified framework for the first time, MUTR adopts a DETR-style transformer and is capable of segmenting video objects designated by either text or audio reference. Specifically, we introduce two strategies to fully explore the temporal relations between videos and multi-modal signals. 
Firstly, for low-level temporal aggregation before the transformer, we enable the multi-modal references to capture multi-scale visual cues from consecutive video frames. This effectively endows the text or audio signals with temporal knowledge and boosts the semantic alignment between modalities.
Secondly, for high-level temporal interaction after the transformer, we conduct inter-frame feature communication for different object embeddings, contributing to better object-wise correspondence for tracking along the video.
On Ref-YouTube-VOS and AVSBench datasets with respective text and audio references, MUTR achieves \textbf{+4.2\%} and \textbf{+8.7\%} $\mathcal{J}\&\mathcal{F}$ improvements to \textit{state-of-the-art} methods, demonstrating our significance for unified multi-modal VOS. Code is released at \textcolor{citecolor}{\url{https://github.com/OpenGVLab/MUTR}}.

\end{abstract}

\section{Introduction}\label{sec:introduction}
Multi-modal video object segmentation (VOS) aims to track and segment particular object instances across the video sequence referred by a given multi-modal signal, including referring video object segmentation (RVOS) with language reference, and audio-visual video object segmentation (AV-VOS) with audio reference.
Different from the vanilla VOS~\cite{xu2018youtube,yan2023panovos} with only visual information, the multi-modal VOS is more challenging and in urgent demand, which requires a comprehensive understanding of different modalities and their temporal correspondence across frames.

There exist two main challenges in multi-modal VOS. Firstly, it requires to not only explore the rich spatial-temporal consistency in a video, but also align the multi-modal semantics among image, language, and audio. Current approaches mainly focus on the visual-language or visual-audio modal fusion within independent frames, simply by cross-modal attention~\cite{chen2019see,hu2020bi,shi2018key} or dynamic convolutions~\cite{margffoy2018dynamic} for feature interaction. This, however, neglects the multi-modal temporal information across frames, which is significant for consistent object segmentation and tracking along the video.
Secondly, for the given references of two modalities, language and audio, existing works adopt different architecture designs and training strategies to separately tackle their modal-specific characteristics. Therefore, a powerful and unified framework for multi-modal VOS still remains an open question.

To address these challenges, we propose \textbf{MUTR}, a \textbf{M}ulti-modal \textbf{U}nified \textbf{T}emporal transformer for \textbf{R}eferring video object segmentation. Our approach, for the first time, presents a generic framework for both language and audio references, and enhances the interaction between temporal frames and multi-modal signals. In detail, we adopt a DETR-like~\cite{carion2020end} encoder-decoder transformer, which serves as the basic architecture to process visual information within different frames. On top of this, we introduce two attention-based modules respectively for low-level multi-modal temporal aggregation (MTA), and high-level multi-object temporal interaction (MTI).
Firstly before the transformer, we utilize the encoded multi-modal references as queries to aggregate informative visual and temporal features via the MTA module. We concatenate the visual features of adjacent frames and adopt sequential attention blocks for multi-modal tokens to progressively capture temporal visual cues of different image scales. This contributes to better low-level cross-modal alignment and temporal consistency.
Then, we regard the multi-modal tokens after MTA as object queries and feed them into the transformer for frame-wise decoding.
After that, we apply the MTI module to conduct inter-frame object-wise interaction, and maintain a set of video-wise query representations for associating objects across frames inspired by~\cite{heo2022vita}. Such a module enhances the instance-level temporal communication and benefits the visual correspondence for segmenting the same object in a video. Finally, we utilize a segmentation head following previous works~\cite{wu2022language,wu2021seqformer} to output the final object mask referred by multi-modality input.

To evaluate our effectiveness, we conduct extensive experiments on several popular benchmarks for multi-modal VOS.
RVOS with language reference (Ref-YouTube-VOS~\cite{seo2020urvos} and Ref-DAVIS 2017~\cite{khoreva2019video}), and one benchmark for AV-VOS with audio reference (AVSBench~\cite{zhou2022audio}). 
On Ref-YouTube-VOS~\cite{seo2020urvos} and Ref-DAVIS 2017~\cite{khoreva2019video} with language references, MUTR surpasses the state-of-the-art method ReferFromer~\cite{wu2022language} by +4.2\%  and +4.1\% $\mathcal{J}\&\mathcal{F}$ scores, respectively. On AV-VOS~\cite{zhou2022audio} with audio references, we also outperform Baseline~\cite{zhou2022audio} by +8.7\% $\mathcal{J}\&\mathcal{F}$ score.

Overall, our contributions are summarized as follows:

\begin{itemize}
   \item For the first time, we present a unified transformer architecture, MUTR, to tackle video object segmentation referred by multi-modal inputs, i.e., language and audio.

   \item To better align the temporal information with multi-modal signals, we propose two attention-based modules, MTA and MTI, respectively for low-level multi-scale aggregation and high-level multi-object interaction, achieving superior cross-modal understanding in a video.

   \item On benchmarks of two modalities, our approach both achieves state-of-the-art results, e.g., 
   +4.2 \% and +4.1\% $\mathcal{J}\&\mathcal{F}$ for Ref-YouTube-VOS  and  Ref-DAVIS 2017, +8.7\% $\mathcal{J}\&\mathcal{F}$ for AV-VOS. This fully indicates the significance and generalization ability of MUTR.

\end{itemize}

\section{Related Work}
\paragraph{Referring video object segmentation (R-VOS).} R-VOS introduces the language expression for target object tracking and segmentation, following the trend of vision-language learning~\cite{zhang2022tip,zhang2023prompt,zhu2023not,fang2023instructseq}.
Existing R-VOS methods can be broadly classified into three categories. One of the most straightforward ideas is to apply referring image segmentation methods~\cite{ding2021vision, yang2022lavt, wang2022cris} independently to video frames, such as RefVOS~\cite{bellver2020refvos}. Obviously, it disregards the temporal information, which makes it difficult to process common video challenges like object disappearance in reproduction.
Another approach involves propagating the target mask detected from key frame and selecting the object to be segmented based on a visual grounding model~\cite{kamath2021mdetr,luo2020multi}. Although it applies the temporal information to some extent, its complex multi-stage training approach is not desirable. The recent work MTTR~\cite{botach2022end} and ReferFormer~\cite{wu2022language} have employed query-based mechanisms. Nevertheless, they are end-to-end frameworks, they perform R-VOS task utilizing image-level segmentation. Constrastly, our unified framework fully explores video-level visual-attended language information for low-level temporal aggregation.

\paragraph{Audio-visual video object segmentation (AV-VOS).} Inspired by recent multi-modality efforts~\cite{zhang2023llama,gao2023llama,lin2023sphinx,wang2023mathcoder,guo2023point,han2023imagebind,han2023onellm}, AV-VOS is proposed for predicting pixel-level individual positions based on a given sound signal. There is little previous work on audio-visual video object segmentation. Until recently~\cite{zhou2022audio} proposed the audio-visual video object segmentation dataset. Different from it,~\cite{mo2023av} is based on the recent visual foundation model Segment Anything Model~\cite{kirillov2023segment,zhang2023personalize} to achieve audio-visual segmentation. However, all of them lack the temporal alignment between multi-modal information.

\begin{figure*}[t!]
  \centering
    \includegraphics[width=\textwidth]{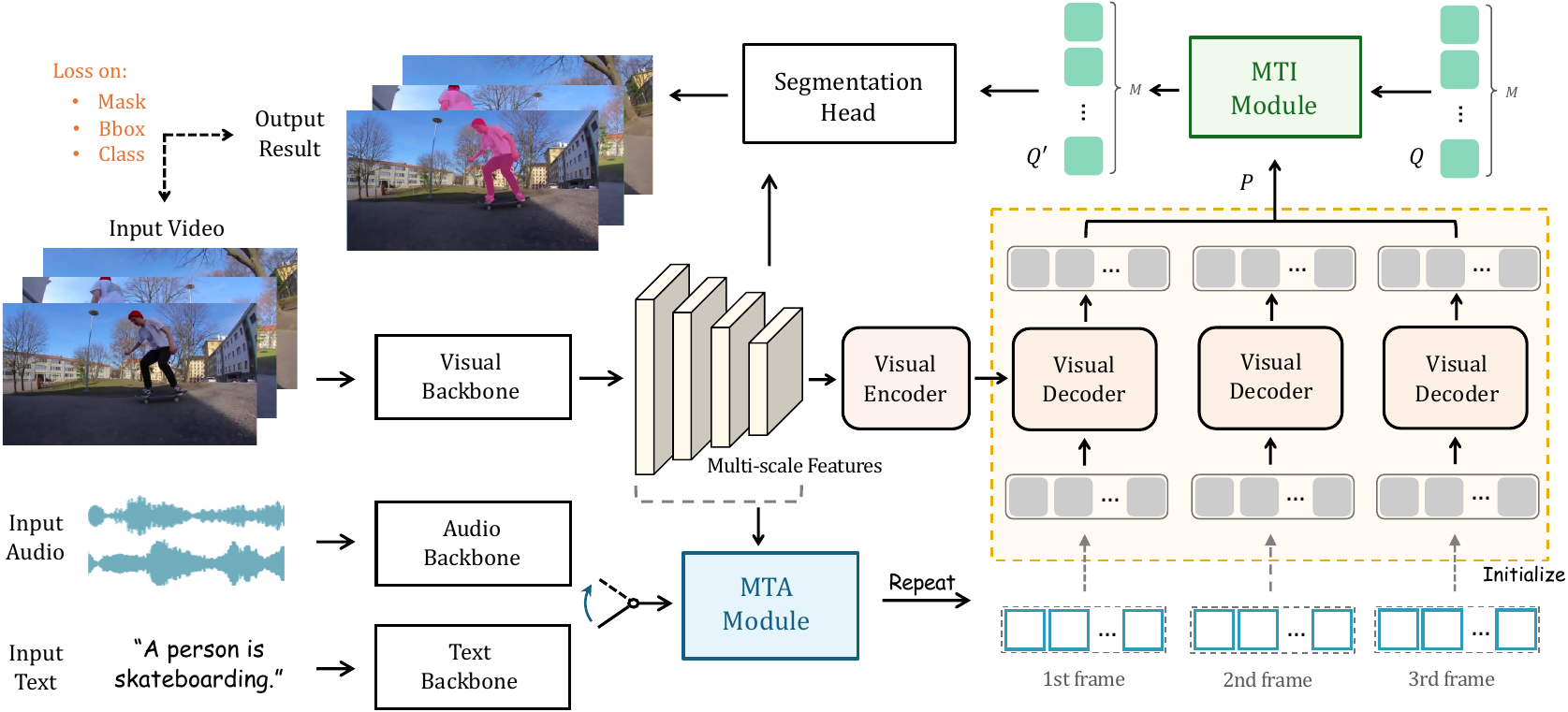}
    \vspace{0.05cm}
   \caption{\textbf{The Overall Pipeline of MUTR for referring video object segmentation.} We present a unified transformer architecture to tackle video object segmentation referred by multi-modal inputs. We propose MTA module and MTI module for low-level multi-scale aggregation and high-level multi-object interaction, respectively.}
    \label{pipeline}
\end{figure*}

\section{Method}
In this section, we illustrate the details of our MUTR for multi-modal video object segmentation. We first describe the overall pipeline in Section~\ref{overall}. Then, in Section~\ref{mta} and Section~\ref{mti}, we respectively
elaborate on the proposed designs of the multi-scale temporal aggregation module (MTA), and multi-object temporal interaction module (MTI).

\subsection{Overall Pipeline}
\label{overall}
The overall pipeline of MUTR is shown in Figure~\ref{pipeline}. We adopt a DETR-based~\cite{carion2020end} transformer as our basic architecture, including a visual backbone, a visual encoder and a decoder, on top of which, two modules MTA and MTI are proposed for temporal multi-modal interaction. In this section, we successively introduce the pipeline of MUTR for video object segmentation.

\paragraph{Feature Backbone.}
Given an input video-text/audio pair, we first sample $T$ frames from the video clip, and utilize the visual backbone and a pre-trained text/audio backbone to extract the image and multi-modal features. Specifically, we utilize ResNet~\cite{he2016deep} or Swin Transformer~\cite{liu2021swin} as the visual backbone, and obtain the multi-scale visual features of the $2^{nd}, 3^{rd}, 4^{th}$ stages. Concurrently, for the text reference, we employ an off-the-shelf language model, RoBERTa~\cite{liu2019roberta}, to encode the linguistic embedding tokens. For the audio reference, we first process it as a spectrogram transform via a short-time Fourier Transform and then feed it into a pre-trained VGGish~\cite{hershey2017cnn} model. After the text/audio encoding, a linear projection layer is adopted to align the multi-modal feature dimension with the visual features. Note that, following previous work~\cite{wu2022language}, we adopt an early fusion module in the visual backbone to inject preliminary text/audio knowledge into visual features. 

\paragraph{MTA Module.}
On top of feature extraction, we feed the visual and text/audio features into the multi-scale temporal aggregation module (MTA). We concatenate the visual features of adjacent frames, and adopt cascaded cross-attention blocks to enhance the multi-scale and multi-modal feature fusion, which is specifically described in Section~\ref{mta}.

\paragraph{Visual Encoder-decoder Transformer.}
The basic transformer consists of a visual encoder and a visual decoder, which processes the video in a frame-independent manner to focus on the feature fusion within a single frame. In detail, the visual encoder adopts vanilla self-attention blocks to encode the multi-scale visual features. 
The visual decoder regards the encoded visual features as the key and value, and the output references from the MTA module as learnable object queries for decoding. 
Unlike the randomly initialized queries in traditional DETR~\cite{carion2020end}, ours are input-conditioned ones obtained via MTA module, which contains video-level multi-modal prior knowledge. With the visual decoder, the object queries gain rich instance information, which provides effective cues for the final segmentation process.

\begin{figure*}[t!]
\begin{minipage}[t]{0.52\textwidth}
\includegraphics[width=\textwidth]{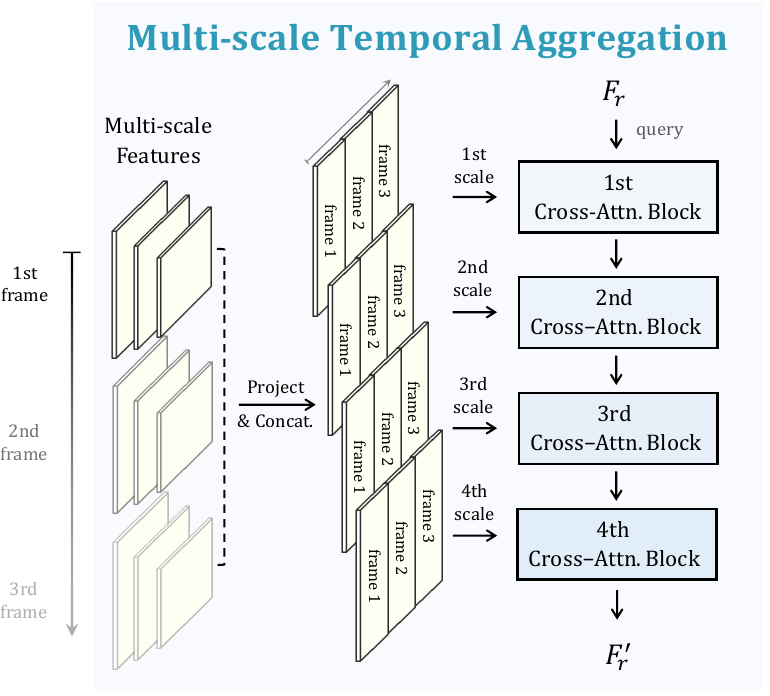}
\caption{\textbf{Multi-scale Temporal Aggregation.} For low-level multi-modal temporal aggregation, we propose MTA module for inter-frame interaction, which generates tokens with multi-modal knowledge as the input queries for transformer decoding.}
\label{fig_mta}
\end{minipage} 
\hspace{0.3cm}
\begin{minipage}[t]{0.42\textwidth}
\includegraphics[width=\textwidth]{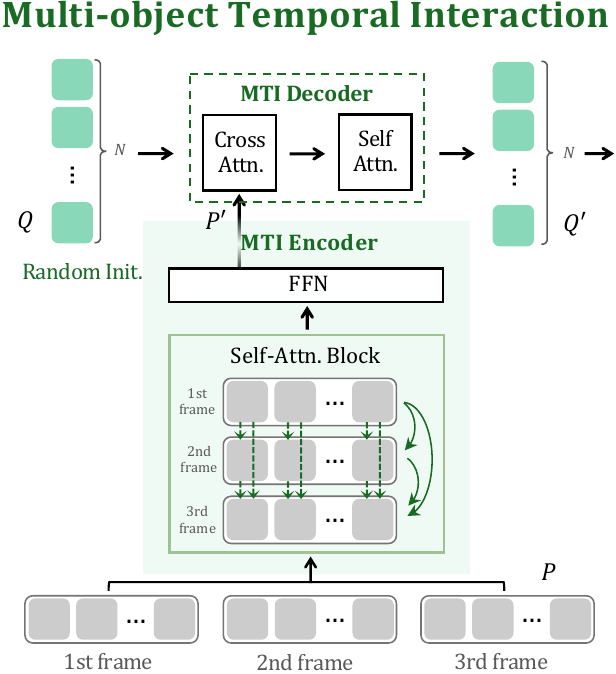}
\caption{\textbf{Multi-object Temporal Interaction.} We introduce MTI module for inter-frame object-wise interaction, and maintain a set of video-wise query representations for associating objects across frames.}
\label{fig_mti}
\end{minipage}
\end{figure*}

\paragraph{MTI Module.}
After the visual transformer, a multi-object temporal interaction (MTI) module is proposed for object-wise interaction, which is described in Section~\ref{mti}. In detail, we utilize an MTI encoder to communicate temporal features of the same object in different views. Then an MTI decoder is proposed to grasp information into a set of video-wise query representations for associating objects across frames, inspired by~\cite{heo2022vita}. 

\paragraph{Segmentation Head and Loss Function.}
On top of the components introduced above, we obtain the final mask predictions from the extracted multi-modal features via a segmentation head. We follow previous works~\cite{wu2022language,wu2021seqformer} to design the segmentation head that contains a bounding box head, a classification head, and a mask head. Then, we find the best assignment from the predictions of MUTR by using Hungarian Matching~\cite{carion2020end}. During training, we calculate three losses in MUTR, which are focal loss~\cite{lin2017focal} $\mathcal{L}_{c l s}$ on the predictions of referred object sequence, $\mathcal{L}_{\text {box }}$ on the bounding box of predicted instance, and $\mathcal{L}_{\text {mask }}$ on the predicted object masks. In detail, $\mathcal{L}_{\text {box }}$ is the combination of $L_1$ loss and GIoU loss~\cite{rezatofighi2019generalized}, and $\mathcal{L}_{\text {mask }}$ is the summation of the Dice~\cite{milletari2016v} and binary focal loss. The whole loss function is formulated as
\begin{align}
\label{loss}
\begin{split}
    & \mathcal{L} = \lambda_{cls}\ \mathcal{L}_{cls} + \lambda_{box}\ \mathcal{L}_{box} + \lambda_{mask}\ \mathcal{L}_{mask}\ ,
\end{split}
\end{align}
where $\lambda_{cls}$, $\lambda_{box}$ and $\lambda_{mask}$ denote the weights for $\mathcal{L}_{cls}$, $\mathcal{L}_{box}$ and $\mathcal{L}_{mask}$. 

\subsection{Multi-scale Temporal Aggregation}
\label{mta}
To boost both the multi-modal and multi-frame feature fusion, we introduce \textbf{M}ulti-scale \textbf{T}emporal \textbf{A}ggregation module for low-level temporal aggregation. The proposed MTA module generates a set of object queries that contain multi-modal knowledge for subsequent transformer decoding.

\paragraph{Multi-scale Temporal Transform.}
As shown in Figure~\ref{fig_mta}, the MTA module take the text/audio features $F_r$, and multi-scale visual features as input, i.e., the extracted features of $2^{nd}, 3^{rd}, 4^{th}$ stages from the visual backbone. We first utilize linear projection layers on the multi-scale features to transform them into the same dimension. Specifically, we separately utilize $1\times1$ convolution layers on the $2^{nd},3^{rd},4^{th}$ scale features, and an additional $3\times3$ convolution layer on the $4^{th}$ stage features to obtain the $5^{th}$ scale features. We denote the projected features as $\{F_{vj}^i\}$, where $2\le i \le 5,\ 1 \le j\le T$ represent the stage number and frame number. After that, we concatenate the visual features of adjacent frames for each scale, formulated as
\begin{align}
\label{mta_concat}
\begin{split}
    & F_{v}^i = \operatorname{Concat}(F_{v1}^i,\ F_{v2}^i,\ ...,\ F_{vj}^i,\ ...,\ F_{vT}^i), \\
\end{split}
\end{align} 
where $2\le i \le 5,\ 1 \le j\le T$, $F_{vj}^i$ represents the projected $j^{th}$ frame features of $i^{th}$ scale, and $\{F_{v}^i\}_{i=2}^{5}$ is the final transformed multi-scale visual feature. 
Then, the resulting multi-modal temporal features are regarded as the key and value in the following cross-attention blocks.

\paragraph{Multi-modal Cross-attention.}
On top of this, we adopt sequential cross-attention mechanisms for multi-modal tokens to progressively capture temporal visual cues of different image scales. 
We adopt four cross-attention blocks that are assigned to each scale respectively for multi-scale temporal feature extracting.
In each attention block, the text/audio features serve as the query, while the multi-scale visual features serve as the key and value. We formulate it as
\begin{align}
\label{mta_eq}
\begin{split}
    & F_f = \operatorname{Block}_{i-1}(F_r,\ F_{v}^i,\ F_{v}^i),\ 2 \le i \le 5, \\
\end{split}
\end{align}
where $\operatorname{Block}$ represents the sequential cross-attention blocks in MTA module, $F_f$ is the output multi-modal tokens that contain the multi-modal information. 

After that, we simply repeat the class token of $F_f$ for $T\times N$ times, where $T$ is the frame number and $N$ is the query number. We adopt them as the initialized queries fed into the visual transformer for frame-wise decoding. With the MTA module, the pre-initialized input queries obtain prior multi-scale knowledge and temporal information for better multi-modal alignment during subsequent decoding.

\subsection{Multi-object Temporal Interaction}
\label{mti}
As the visual transformer adopts a frame-independent manner and fails to interact information among multiple frames, we further introduce a \textbf{M}ulti-object \textbf{T}emporal \textbf{I}nteraction module to conduct inter-frame object-wise interaction. This module enhances the high-level temporal communication of objects, and benefits the visual correspondence for effective segmentation. The details of MTI are shown in Figure~\ref{fig_mti}, which consists of an MTI encoder and an MTI decoder.

\paragraph{MTI Encoder.}
We obtain the object query outputs $P$ of each frame from the transformer decoder, and feed them into the MTI encoder, which contains a self-attention layer to conduct object-wise interaction across multiple frames, and a feed-forward network layer for feature transformation. To achieve more efficient implementation, we adopt shifted window-attention~\cite{liu2021swin} with linear computational complexity in the self-attention layer. The process of MTI encoder is formulated as
\begin{align}
\label{mti_encoder}
\begin{split}
    & P' = \operatorname{MTI\_Encoder}(P)\\
\end{split}
\end{align}
where $\operatorname{MTI\_Encoder}$ denotes the MTI encoder, and $P'$ is the outputs of MTI encoder.

\begin{table*}[t!]
\vspace{-0.2cm}
\centering
\vspace{0.12cm}
\small
\centering
\setlength\tabcolsep{10pt}
\begin{tabular}{l|c|ccc|ccc}
\toprule[1.1pt]
    \makecell*[c]{\multirow{2}*{Method}} 
    &\makecell*[c]{\multirow{2}*{Backbone}} &  \multicolumn{3}{c|}{Ref-YouTube-VOS}  & \multicolumn{3}{c}{Ref-DAVIS 2017} \\
    & &$\mathcal{J}\&\mathcal{F}$ & $\mathcal{J}$ & $\mathcal{F}$ & $\mathcal{J}\&\mathcal{F}$ & $\mathcal{J}$ & $\mathcal{F}$ \\
\cmidrule(lr){1-8} 
CMSA~\cite{ye2019cross} & \makecell*[c]{\multirow{6}*{ResNet-50}} & 34.9 & 33.3 & 36.5 & 34.7 & 32.2 & 37.2 \vspace{-3pt}\\
URVOS~\cite{seo2020urvos} & & 47.2 & 45.3 & 49.2 & 51.5 & 47.3 & 56.0 \\
LBDT-4~\cite{ding2022language} & & 48.2 & 50.6 & 49.4 & - & - & - \\
YOFO~\cite{li2022you} & & 48.6 & 47.5 & 49.7 & 53.3 & 48.8 & 57.9  \\
ReferFormer~\cite{wu2022language} & & 58.7 & 57.4 & 60.1 & 61.1 & 58.0 & 64.1 \\
\textbf{MUTR} & & \textbf{61.9} & \textbf{60.4} & \textbf{63.4} & \textbf{65.3} & \textbf{62.4} & \textbf{68.2} \\
\cmidrule(lr){1-8}  
CITD~\cite{liang2021rethinking} &\makecell*[c]{\multirow{3}*{ResNet-101}} & 56.4 & 54.8 & 58.1 & - & - & -  \vspace{-3pt}\\
ReferFormer~\cite{wu2022language} & & 59.3 & 58.1 & 60.4 & 61.0 & 58.1 & 63.8 \\
\textbf{MUTR}& & \textbf{63.6} & \textbf{61.8} & \textbf{65.4} & \textbf{65.3} & \textbf{61.9}& \textbf{68.6} \\
\cmidrule(lr){1-8}  
ReferFormer~\cite{wu2022language} &\makecell*[c]{\multirow{2}*{Swin-L}} & 64.2 & 62.3 & 66.2 & 63.9 & 60.8 & 67.0 \vspace{-3pt}\\
\textbf{MUTR}& & \textbf{68.4} & \textbf{66.4} & \textbf{70.4} & \textbf{68.0} & \textbf{64.8} & \textbf{71.3} \\
\cmidrule(lr){1-8}  
MTTR~\cite{botach2022end} &\makecell*[c]{\multirow{4}*{Video-Swin-T}} & 55.3 & 54.0 & 56.6 & - & - & - \vspace{-3pt}\\
MANet~\cite{chen2022multi} & & 55.6 & 54.8 & 56.5 & - & - & - \\
ReferFormer~\cite{wu2022language} & & 62.6 & 59.9 & 63.3 & 62.8 & 60.8 & 67.0 \\
\textbf{MUTR}& & \textbf{64.0} & \textbf{62.2} & \textbf{65.8} & \textbf{66.5} & \textbf{63.0} & \textbf{70.0} \\
\cmidrule(lr){1-8}  
VLT~\cite{ding2022vlt} &\makecell*[c]{\multirow{3}*{Video-Swin-B}} & 63.8 & 61.9 & 65.6 & 61.6 & 58.9 & 64.3 \vspace{-3pt}\\
ReferFormer~\cite{wu2022language} && 64.9 & 62.8 & 67.0 & 64.3 & 60.7 & 68.0 \\
\textbf{MUTR}& & \textbf{67.5} & \textbf{65.4} & \textbf{69.6}& \textbf{66.4} & \textbf{62.8} & \textbf{70.0} \\
\bottomrule[1.1pt]
\end{tabular}
\caption{\textbf{Performance of MUTR on Ref-YouTube-VOS and Ref-DAVIS 2017 Datasets.} We report the results of MUTR and prior works on multiple backbones, where our MUTR shows the \textit{state-of-the-art} performance on all datasets.
}
\label{tab:ref-ytb-davis}
\end{table*}

\paragraph{MTI Decoder.}
Based on the MTI encoder, we maintain a set of video-wise query $Q$ for associating objects across frames, which are randomly initialized. We regard the outputs from MTI encoder as the key and value, and feed them and video-wise queries $Q$ into MTI decoder for video-wise decoding. The MTI decoder consists of a cross-attention layer, a self-attention layer, and a feed-forward network layer. We formulate them as
\begin{align}
\label{mti_decoder}
\begin{split}
    & Q' = \operatorname{MTI\_Decoder}(Q,\ P',\ P')\\
\end{split}
\end{align}
where $\operatorname{MTI\_Decoder}$ represents the MTI decoder, $Q'$ is the outputs of MTI decoder. 
In this way, the proposed MTI module promotes high-level temporal fusion and enhances the connection and interaction of the same objects in different frames, which further contributes to effective segmentation.

\subsection{Joint Training for Multi-modality}

As a unified VOS framework for multi-modality, MUTR has the potential to segment video objects referred by either text or audio reference. To achieve this, we conduct joint training by combining both text- and audio-referred datasets. Specifically, to balance the data amount of two modalities, the joint training data is composed of partial Ref-YouTube-VOS~\cite{seo2020urvos} (text reference) and the entire AVSBench S4~\cite{zhou2022audio} (audio reference). We sample a subset of Ref-YouTube-VOS for training (10,093 clips (5 frames per clip) out of 72,920), for which we utilize only one description for videos with multiple text descriptions, and filter out half of the instances based on odd-index positions for training.

For text or audio reference, we accordingly switch to their respective encoders for feature encoding, i.e., RoBERTa for text and VGGish for audio. Then, they share the same subsequent network modules, including the MTA, visual encoder, visual decoder, MTI, and segments head. By our proposed temporal and cross-modality interaction modules, the jointly trained MUTR can obtain superior performance on either of the two modalities.

\section{Experiments}

\begin{table*}[t!]
\vspace{-0.2cm}
\centering
\vspace{0.12cm}
\small
\centering
\setlength\tabcolsep{10pt}
\begin{tabular}{l|c|ccc|ccc}
\toprule[1.1pt]
    \makecell*[c]{\multirow{2}*{Method}} 
    &\makecell*[c]{\multirow{2}*{Backbone}} &  \multicolumn{3}{c|}{AVSBench S4}  & \multicolumn{3}{c}{AVSBench MS3} \\
    & &$\mathcal{J}\&\mathcal{F}$ & $\mathcal{J}$ & $\mathcal{F}$ & $\mathcal{J}\&\mathcal{F}$ & $\mathcal{J}$ & $\mathcal{F}$ \\
\cmidrule(lr){1-8} 
LVS~\cite{chen2021localizing} & ResNet-18 & 44.5  & 37.9 & 51.0  & 31.3 &29.5 &33.0  \\
SST~\cite{duke2021sstvos}  & ResNet-50 &73.2  & 66.3 & 80.1 & 49.9 & 42.6& 57.2\\
LGVT~\cite{zhang2021learning} & Swin-B &{81.1}  & {74.9} & {87.3}  & 50.0 & 40.7 &59.3 \\
Baseline~\cite{zhou2022audio} &ResNet-50   &{78.8}  & {72.8} & {84.8} & 52.9 & 47.9 &57.8 \\
Baseline~\cite{zhou2022audio} &PvT-V2   &{83.3}  & {78.7} & {87.9} & 59.3 & 54.0 & 64.5\\
\cmidrule(lr){1-8}
\multirow{7}*{MUTR}  & ResNet-50 &{83.0}  & {78.6} & {87.3}  & 61.6 & 57.0 & 66.1\\
& ResNet-101 &{83.1}  & {78.5} & {87.6}  & 63.7 & 59.0 & 68.3\\
& PvT-V2 &{85.1}  & {80.7} & {89.5} & 67.9 & 63.7 & 72.0 \\
& Swin-L &{85.7}  & {81.5} & {89.8} & 69.0 & 65.0 & 73.0 \\
& Video-Swin-T &{83.0}  & {78.7} & {87.2} & 64.0 & 59.2 & 68.7  \\
& Video-Swin-S &{84.1}  & {79.8} & {88.3} & 67.3 & 62.7 & 71.8\\
& Video-Swin-B &{85.7}  & {81.6} & {89.7} & 68.8 & 64.0 & 73.5 \\
\bottomrule[1.1pt]
\end{tabular}

\caption{\textbf{Performance of MUTR on AVSBench Dataset.} MUTR surpasses the \textit{state-of-the-art} method.
}
\label{tab:avs}
\vspace{0.1cm}
\end{table*}

\begin{figure*}
\vspace{0.1cm}
\centering
\begin{minipage}[t!]{0.48\linewidth}
\begin{adjustbox}{width=\linewidth}
\begin{tabular}{l|c c c c c c}
    \toprule
         Methods   & $\mathcal{J}\&\mathcal{F}$ & $\mathcal{J}$ &  $\mathcal{F}$   \\ 
        \cmidrule(lr){1-1} \cmidrule(lr){2-4}
        ReferFormer~\cite{wu2022language} & 32.5& 32.6& 32.4 \\
        \textbf{MUTR}$^{*}$ &39.9 & 39.4 & 40.5 \\
        \rowcolor{gray!10}\textbf{MUTR} & 41.3& 40.6  &42.0 \\
    \bottomrule
\end{tabular}
\end{adjustbox}
\tabcaption{Performance of MUTR on Ref-YouTube-VOS by \textbf{Multi-modality Joint Training.}}
\label{tab:joint_text}
\end{minipage}\quad\quad
\begin{minipage}[t!]{0.47\linewidth}
\centering
\begin{adjustbox}{width=\linewidth}
\begin{tabular}{l|c c c c c c}
    \toprule
         Methods   & $\mathcal{J}\&\mathcal{F}$ & $\mathcal{J}$ &  $\mathcal{F}$   \\ 
        \cmidrule(lr){1-1} \cmidrule(lr){2-4}
        Baseline~\cite{zhou2022audio} &78.8 & 72.8 & 84.8\\
        \textbf{MUTR}$^{*}$ &79.7 & 74.5 & 84.9 \\
        \rowcolor{gray!10}\textbf{MUTR} & 81.4& 76.8  &85.9 \\
    \bottomrule
\end{tabular}
\end{adjustbox}
\tabcaption{Performance of MUTR on AVSBench S4 by \textbf{Multi-modality Joint Training.}}
\label{tab:joint_audio}
\end{minipage}
\vspace{0.1cm}
\end{figure*}

\begin{figure*}[t!]
\begin{minipage}[t!]{0.484\linewidth}
\centering
\vspace{0.1cm}
\begin{adjustbox}{width=\linewidth}
\begin{tabular}{cccccc}
    \toprule
    \multicolumn{2}{c}{Components} &\makecell*[c]{\multirow{2}*{\shortstack{\vspace*{1.2pt}\\Block\\\vspace*{0.3pt}\\Num.}}} 
    &\makecell*[c]{\multirow{2}*{$\mathcal{J\&F}$}} 
    &\makecell*[c]{\multirow{2}*{$\mathcal{J}$}}
    &\makecell*[c]{\multirow{2}*{$\mathcal{F}$}} \\
    \cmidrule(lr){1-2} 
      Multi-scale &Temporal &&&&\\
     \cmidrule(lr){1-1}  \cmidrule(lr){2-2}  \cmidrule(lr){3-3}  \cmidrule(lr){4-4}  \cmidrule(lr){5-5}  \cmidrule(lr){6-6}  
     \checkmark &- &1 &61.3 &59.7 &62.7\\
     - &\checkmark &1 &60.4 &58.9 &61.9\\
     \rowcolor{gray!10}\checkmark &\checkmark &1 &\textbf{61.9} &\textbf{60.4} &\textbf{63.4}\\
     \checkmark &\checkmark &2 &60.7 &59.3 &62.2\\
     \checkmark &\checkmark &3 &60.4 &59.1 &61.7\\
    \bottomrule
\end{tabular}
\end{adjustbox}
\tabcaption{Ablation Study of MTA Module.}
\label{tab:mta}
\end{minipage}\qquad
\begin{minipage}[t!]{0.451\linewidth}
\centering
\small
\vspace{0.1cm}
\begin{adjustbox}{width=\linewidth}
\begin{tabular}{cccccc}
    \toprule
    \multicolumn{2}{c}{Components} &\makecell*[c]{\multirow{2}*{\shortstack{\vspace*{1.2pt}\\Block\\\vspace*{0.3pt}\\Num.}}} 
    &\makecell*[c]{\multirow{2}*{$\mathcal{J\&F}$}} 
    &\makecell*[c]{\multirow{2}*{$\mathcal{J}$}}
    &\makecell*[c]{\multirow{2}*{$\mathcal{F}$}} \\
    \cmidrule(lr){1-2} 
      Encoder &Decoder &&&&\\
     \cmidrule(lr){1-1}  \cmidrule(lr){2-2}  \cmidrule(lr){3-3}  \cmidrule(lr){4-4}  \cmidrule(lr){5-5}  \cmidrule(lr){6-6}  
     \checkmark &- &3 &60.3 &58.8 &61.9\\
     - &\checkmark &3 &61.2 &60.0 &62.6\\
     \rowcolor{gray!10}\checkmark &\checkmark &3 &\textbf{61.9} &\textbf{60.4} &\textbf{63.4}\\
     \checkmark &\checkmark &2 &61.1 &59.5 &62.6\\
     \checkmark &\checkmark &1 &60.8 &59.3 &62.3\\
    \bottomrule
\end{tabular}
\end{adjustbox}
\tabcaption{Ablation Study of MTI Module.}
\label{tab:mti}
\end{minipage}
\end{figure*}

\subsection{Quantitative Results}\label{s4.3}

\paragraph{Ref-YouTube-VOS.} 

As shown in Table~\ref{tab:ref-ytb-davis}, MUTR outperforms the previous state-of-the-art methods by a large margin under on all datasets. On Ref-YouTube-VOS, MUTR with a lightweight backbone ResNet-50 achieves the superior performance with overall $\mathcal{J}\&\mathcal{F}$ of 61.9\%, an improvement of +3.2\% than the previous state-of-the-art method Referformer. 
By adopting a more powerful backbone Swin-Transformer~\cite{liu2021swin}, MUTR improves the performance to $\mathcal{J}\&\mathcal{F}$ 68.4\%, which is +4.2\% than the previous method ReferFormer~\cite{wu2022language}. Using a more strong backbone, our method has a higher percentage of improvement, which better reflects the robustness of our method on the scaled-up model size. 
To reflect the powerful temporal modeling capability of MUTR, we therefore adopt the video Swin transformer~\cite{liu2022video} as the backbone, which is a spatial-temporal encoder that can effectively capture the spatial and temporal cues simultaneously, to compensate for the temporal limitations of the ReferFormer as discussed in~\cite{hu20221st}. It can be observed that our method significantly outperforms the ReferFormer, which demonstrates the effectiveness of the temporal consistency in our model.

\paragraph{Ref-DAVIS 2017.} On the Ref-DAVIS 2017, our method also achieves the best results under the same backbone setting. Since ReferFormer~\cite{wu2022language} does not include the resultson Ref-DAVIS 2017, we report its results using the official pre-trained models provided by ReferFormer.

\paragraph{AV-VOS.} Table~\ref{tab:avs} shows the performance of our MUTR on the AVSBench dataset. MUTR significantly surpasses all the previous best competitors  ($\mathcal{J}\&\mathcal{F}$ \textbf{83.0\% VS 78.8\%; 61.6\% VS 52.9\%}) with the same ResNet-50 backbone. We also achieve a new state-of-the-art performance with Swin-L~\cite{liu2021swin} backbone.
By employing a stronger backbone, we observe consistent performance improvement of MUTR, indicating the strong generalization of our approach.


\paragraph{Joint Training Datasets.}
We keep most training hyperparameters consistent with our previous text-referred video object segmentation experiments, and adopt ResNet-50 as the visual backbone.
Table~\ref{tab:joint_text} and \ref{tab:joint_audio} present the performance of MUTR by joint training on Ref-YouTube-VOS and AVSBench S4, respectively. Therein, ReferFormer, the `Baseline', and MUTR$^{*}$ are all trained exclusively on text- or audio-referred dataset, while MUTR is trained on the multi-modality joint dataset. As shown, the single unified MUTR by joint training can achieve even better performance than their separate training. This indicates the effectiveness of our proposed architecture to serve as a unified framework simultaneously for text and audio input.

\begin{figure*}[t!]
  \centering
    \includegraphics[width=\textwidth]{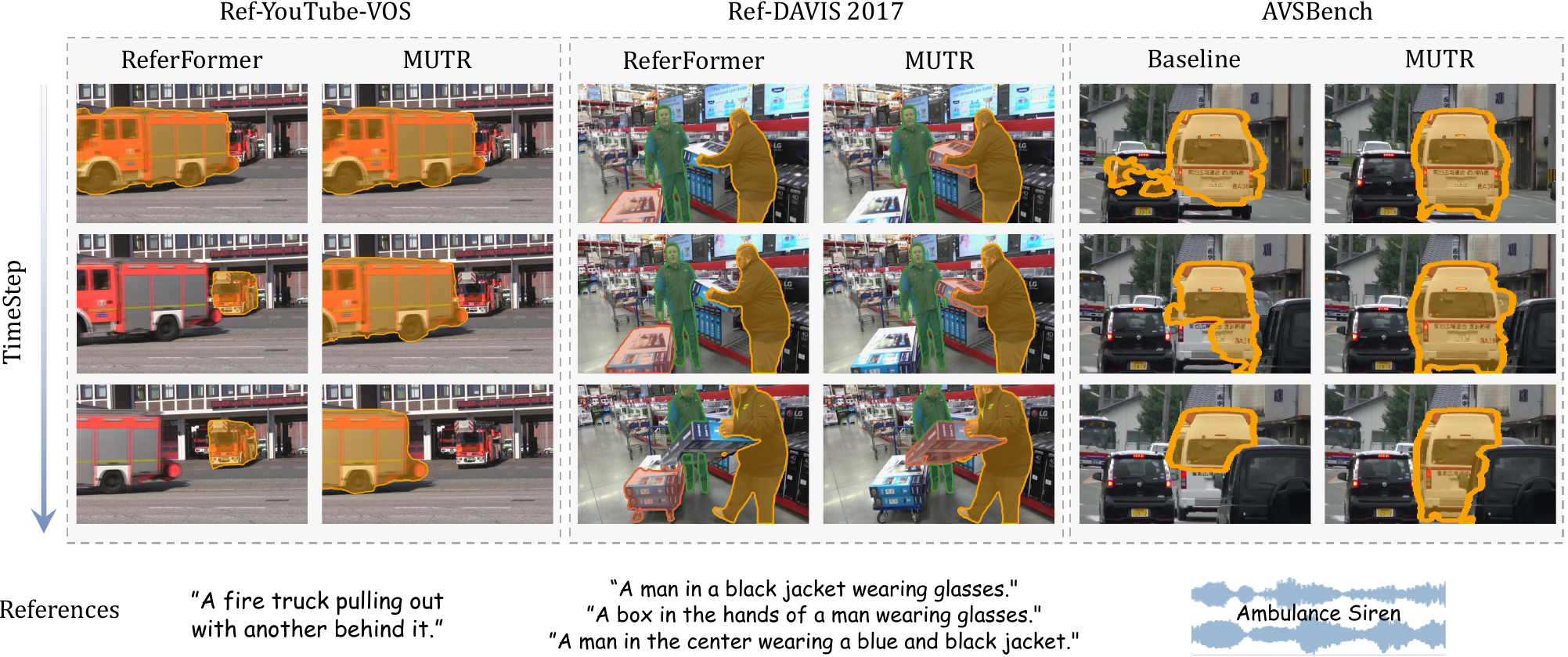}
   \caption{\textbf{Qualitative Results of MUTR.} We visualize the results between ReferFormer~\cite{wu2022language} and MUTR on R-VOS benchmarks and between Baseline~\cite{zhou2022audio} and MUTR on AV-VOS benchmark. Compared with ReferFormer, MUTR performs better on temporal consistency when segmenting multiple similar objects, i.e., fire truck in Ref-YouTube-VOS and box in Ref-DAVIS 2017. Also, compared with the baseline of AV-VOS~\cite{zhou2022audio} that denoted as `Baseline' in this figure, MUTR can handle serve occlusion.}
    \label{qualitative results}
\end{figure*}

\subsection{Qualitative Results}\label{s4.4}

The first two columns of Figure~\ref{qualitative results} visualize some qualitative results in comparison with ReferFormer~\cite{wu2022language}, which lacks inter-frame interaction in terms of temporal dimension. As demonstrated, along with multiple highly similar objects in the video, ReferFormer~\cite{wu2022language} is easier to misidentifies them. In contrast, our MUTR is able to associate all the objects in temporal, which can better track and segment all targets accurately.

The last column of Figure~\ref{qualitative results} visualizes the audio-visual result compared with Baseline~\cite{zhou2022audio} on AVSBbench S4 dataset. With temporal consistency, MUTR can successfully track and segment challenging situations that are surrounded or occluded by similar instances.

\begin{table}[t!]
\begin{adjustbox}{width=\linewidth}
\centering
\begin{tabular}{c c|c c c c c c}
    \toprule
         MTA  & MTI  & $\mathcal{J}\&\mathcal{F}$ & $\mathcal{J}$ &  $\mathcal{F}$  & FPS  &Parameters   \\ 
        \cmidrule(lr){1-1} \cmidrule(lr){2-2} \cmidrule(lr){3-7}
        - &- & 60.2 & 58.7 & 61.7 & 19.64  & 168.1M\\
        - & \checkmark & 60.8 & 59.3 & 62.2 & 19.53  & 176.4M\\
        \checkmark &- & 61.5 & 60.1 & 63.0 & 19.44  & 169.3M\\
        \rowcolor{gray!10}\checkmark &\checkmark & \textbf{61.9} & \textbf{60.4} & \textbf{63.4} & 19.37  & 177.6M\\
    \bottomrule
\end{tabular}
\end{adjustbox}
\tabcaption{\textbf{Ablation Study of the MTA and MTI Modules.}}
\label{tab:mta_mti}
\vspace{-0.15cm}
\end{table}
\subsection{Ablation Studies}\label{s4.5}
In this section, we perform experiments to analyze the main components and hyper-parameters of MUTR. All the experiments are conducted with the ResNet-50 backbone and evaluate their impact by the Ref-YouTube-VOS performance.

\paragraph{Effectiveness of Main Componenets.} Table~\ref{tab:mta_mti} demonstrates the effectiveness of MTA and MTI proposed in our framework.  The performance will be seriously degraded from  61.9\% to 60.2\% by removing MTA and MTI modules. Besides, our MTA and MTI modules introduce a marginal increase in inference latency, demonstrating favorable implementation and parameter efficiency.

\paragraph{Ablation Study on MTA.}
In Table~\ref{tab:mta}, if either the single-scale temporal aggregation or multi-scale aggregation at the image level are adopted, the performance of MUTR would significantly drop to 60.4\% and 61.3\%, respectively, which demonstrates the necessity of the MTA module. We also ablate the number of MTA blocks. As seen in Table~\ref{tab:mta}, more MTA blocks cannot bring further performance improvement, since (1) not enough videos for training; (2) the embedding space of visual and reference is only 256-dimensional, which is difficult to optimize so many parameters.

\paragraph{Ablation Study on MTI.}
As shown in Table~\ref{tab:mti}, the performance of MUTR is improved by using more MTI blocks. A possible reason is that the larger the MTI blocks, the more sufficient temporal communication between instance-level can be performed.
Moreover, using only the encoder or decoder, the performance of MUTR would both decline. 


\section{Conclusion}
This paper proposes a MUTR, a \textbf{M}ulti-modal \textbf{U}nified \textbf{T}emporal transformer for \textbf{R}eferring video object segmentation. A simple yet and effective Multi-scale Temporal Aggregation (MTA) is introduced for multi-modal references to explore low-level multi-scale visual information in video-level. Besides, the high-level Multi-object Temporal Interaction (MTI) is designed for inter-frame feature communication to achieve temporal correspondence between the instance-level across the entire video. Aided by the MTA and MTI, our MUTR achieves new state-of-the-art performance on three R-VOS/AV-VOS benchmarks compared to previous solutions.
We hope the MTA and MTI will help ease the future study of multi-modal VOS and related tasks (e.g., referring video object tracking and video instance segmentation). We do not foresee negative social impact from the proposed work.

\section{Acknowledgement}
This work was supported by National Natural Science Foundation of China (Grant No. 62206272). This work was supported in part by Scientific and Technological Innovation Action Plan of Shanghai Science and Technology
Committee (No.22511101502).


\bibliography{aaai24}

\begin{thebibliography}{52}
\providecommand{\natexlab}[1]{#1}

\bibitem[{Bellver et~al.(2020)Bellver, Ventura, Silberer, Kazakos, Torres, and Giro-i Nieto}]{bellver2020refvos}
Bellver, M.; Ventura, C.; Silberer, C.; Kazakos, I.; Torres, J.; and Giro-i Nieto, X. 2020.
\newblock Refvos: A closer look at referring expressions for video object segmentation.
\newblock \emph{arXiv preprint arXiv:2010.00263}.

\bibitem[{Botach, Zheltonozhskii, and Baskin(2022)}]{botach2022end}
Botach, A.; Zheltonozhskii, E.; and Baskin, C. 2022.
\newblock End-to-end referring video object segmentation with multimodal transformers.
\newblock In \emph{Proceedings of the IEEE/CVF Conference on Computer Vision and Pattern Recognition}, 4985--4995.

\bibitem[{Carion et~al.(2020)Carion, Massa, Synnaeve, Usunier, Kirillov, and Zagoruyko}]{carion2020end}
Carion, N.; Massa, F.; Synnaeve, G.; Usunier, N.; Kirillov, A.; and Zagoruyko, S. 2020.
\newblock End-to-end object detection with transformers.
\newblock In \emph{Computer Vision--ECCV 2020: 16th European Conference, Glasgow, UK, August 23--28, 2020, Proceedings, Part I 16}, 213--229. Springer.

\bibitem[{Chen et~al.(2019)Chen, Jia, Lo, Chen, and Liu}]{chen2019see}
Chen, D.-J.; Jia, S.; Lo, Y.-C.; Chen, H.-T.; and Liu, T.-L. 2019.
\newblock See-through-text grouping for referring image segmentation.
\newblock In \emph{Proceedings of the IEEE/CVF International Conference on Computer Vision}, 7454--7463.

\bibitem[{Chen et~al.(2021)Chen, Xie, Afouras, Nagrani, Vedaldi, and Zisserman}]{chen2021localizing}
Chen, H.; Xie, W.; Afouras, T.; Nagrani, A.; Vedaldi, A.; and Zisserman, A. 2021.
\newblock Localizing visual sounds the hard way.
\newblock In \emph{Proceedings of the IEEE/CVF Conference on Computer Vision and Pattern Recognition}, 16867--16876.

\bibitem[{Chen et~al.(2022)Chen, Hong, Qi, Han, Wang, Qing, Huang, and Li}]{chen2022multi}
Chen, W.; Hong, D.; Qi, Y.; Han, Z.; Wang, S.; Qing, L.; Huang, Q.; and Li, G. 2022.
\newblock Multi-Attention Network for Compressed Video Referring Object Segmentation.
\newblock In \emph{Proceedings of the 30th ACM International Conference on Multimedia}, 4416--4425.

\bibitem[{Ding et~al.(2021)Ding, Liu, Wang, and Jiang}]{ding2021vision}
Ding, H.; Liu, C.; Wang, S.; and Jiang, X. 2021.
\newblock Vision-language transformer and query generation for referring segmentation.
\newblock In \emph{Proceedings of the IEEE/CVF International Conference on Computer Vision}, 16321--16330.

\bibitem[{Ding et~al.(2022{\natexlab{a}})Ding, Liu, Wang, and Jiang}]{ding2022vlt}
Ding, H.; Liu, C.; Wang, S.; and Jiang, X. 2022{\natexlab{a}}.
\newblock VLT: Vision-Language Transformer and Query Generation for Referring Segmentation.
\newblock \emph{IEEE Transactions on Pattern Analysis and Machine Intelligence}.

\bibitem[{Ding et~al.(2022{\natexlab{b}})Ding, Hui, Huang, Wei, Han, and Liu}]{ding2022language}
Ding, Z.; Hui, T.; Huang, J.; Wei, X.; Han, J.; and Liu, S. 2022{\natexlab{b}}.
\newblock Language-bridged spatial-temporal interaction for referring video object segmentation.
\newblock In \emph{Proceedings of the IEEE/CVF Conference on Computer Vision and Pattern Recognition}, 4964--4973.

\bibitem[{Duke et~al.(2021)Duke, Ahmed, Wolf, Aarabi, and Taylor}]{duke2021sstvos}
Duke, B.; Ahmed, A.; Wolf, C.; Aarabi, P.; and Taylor, G.~W. 2021.
\newblock Sstvos: Sparse spatiotemporal transformers for video object segmentation.
\newblock In \emph{Proceedings of the IEEE/CVF Conference on Computer Vision and Pattern Recognition}, 5912--5921.

\bibitem[{Fang et~al.(2023)Fang, Yan, Huang, Zhou, Tian, Dai, and Li}]{fang2023instructseq}
Fang, R.; Yan, S.; Huang, Z.; Zhou, J.; Tian, H.; Dai, J.; and Li, H. 2023.
\newblock InstructSeq: Unifying Vision Tasks with Instruction-conditioned Multi-modal Sequence Generation.
\newblock \emph{arXiv preprint arXiv:2311.18835}.

\bibitem[{Gao et~al.(2023)Gao, Han, Zhang, Lin, Geng, Zhou, Zhang, Lu, He, Yue et~al.}]{gao2023llama}
Gao, P.; Han, J.; Zhang, R.; Lin, Z.; Geng, S.; Zhou, A.; Zhang, W.; Lu, P.; He, C.; Yue, X.; et~al. 2023.
\newblock Llama-adapter v2: Parameter-efficient visual instruction model.
\newblock \emph{arXiv preprint arXiv:2304.15010}.

\bibitem[{Guo et~al.(2023)Guo, Zhang, Zhu, Tang, Ma, Han, Chen, Gao, Li, Li et~al.}]{guo2023point}
Guo, Z.; Zhang, R.; Zhu, X.; Tang, Y.; Ma, X.; Han, J.; Chen, K.; Gao, P.; Li, X.; Li, H.; et~al. 2023.
\newblock Point-bind \& point-llm: Aligning point cloud with multi-modality for 3d understanding, generation, and instruction following.
\newblock \emph{arXiv preprint arXiv:2309.00615}.

\bibitem[{Han et~al.(2023{\natexlab{a}})Han, Gong, Zhang, Wang, Zhang, Lin, Qiao, Gao, and Yue}]{han2023onellm}
Han, J.; Gong, K.; Zhang, Y.; Wang, J.; Zhang, K.; Lin, D.; Qiao, Y.; Gao, P.; and Yue, X. 2023{\natexlab{a}}.
\newblock OneLLM: One Framework to Align All Modalities with Language.
\newblock \emph{arXiv preprint arXiv:2312.03700}.

\bibitem[{Han et~al.(2023{\natexlab{b}})Han, Zhang, Shao, Gao, Xu, Xiao, Zhang, Liu, Wen, Guo et~al.}]{han2023imagebind}
Han, J.; Zhang, R.; Shao, W.; Gao, P.; Xu, P.; Xiao, H.; Zhang, K.; Liu, C.; Wen, S.; Guo, Z.; et~al. 2023{\natexlab{b}}.
\newblock Imagebind-llm: Multi-modality instruction tuning.
\newblock \emph{arXiv preprint arXiv:2309.03905}.

\bibitem[{He et~al.(2016)He, Zhang, Ren, and Sun}]{he2016deep}
He, K.; Zhang, X.; Ren, S.; and Sun, J. 2016.
\newblock Deep residual learning for image recognition.
\newblock In \emph{Proceedings of the IEEE conference on computer vision and pattern recognition}, 770--778.

\bibitem[{Heo et~al.(2022)Heo, Hwang, Oh, Lee, and Kim}]{heo2022vita}
Heo, M.; Hwang, S.; Oh, S.~W.; Lee, J.-Y.; and Kim, S.~J. 2022.
\newblock Vita: Video instance segmentation via object token association.
\newblock \emph{arXiv preprint arXiv:2206.04403}.

\bibitem[{Hershey et~al.(2017)Hershey, Chaudhuri, Ellis, Gemmeke, Jansen, Moore, Plakal, Platt, Saurous, Seybold et~al.}]{hershey2017cnn}
Hershey, S.; Chaudhuri, S.; Ellis, D.~P.; Gemmeke, J.~F.; Jansen, A.; Moore, R.~C.; Plakal, M.; Platt, D.; Saurous, R.~A.; Seybold, B.; et~al. 2017.
\newblock CNN architectures for large-scale audio classification.
\newblock In \emph{2017 ieee international conference on acoustics, speech and signal processing (icassp)}, 131--135. IEEE.

\bibitem[{Hu et~al.(2022)Hu, Chen, Gao, Ji, and Bai}]{hu20221st}
Hu, Z.; Chen, B.; Gao, Y.; Ji, Z.; and Bai, J. 2022.
\newblock 1st Place Solution for YouTubeVOS Challenge 2022: Referring Video Object Segmentation.
\newblock \emph{arXiv preprint arXiv:2212.14679}.

\bibitem[{Hu et~al.(2020)Hu, Feng, Sun, Zhang, and Lu}]{hu2020bi}
Hu, Z.; Feng, G.; Sun, J.; Zhang, L.; and Lu, H. 2020.
\newblock Bi-directional relationship inferring network for referring image segmentation.
\newblock In \emph{Proceedings of the IEEE/CVF conference on computer vision and pattern recognition}, 4424--4433.

\bibitem[{Kamath et~al.(2021)Kamath, Singh, LeCun, Synnaeve, Misra, and Carion}]{kamath2021mdetr}
Kamath, A.; Singh, M.; LeCun, Y.; Synnaeve, G.; Misra, I.; and Carion, N. 2021.
\newblock Mdetr-modulated detection for end-to-end multi-modal understanding.
\newblock In \emph{Proceedings of the IEEE/CVF International Conference on Computer Vision}, 1780--1790.

\bibitem[{Khoreva, Rohrbach, and Schiele(2019)}]{khoreva2019video}
Khoreva, A.; Rohrbach, A.; and Schiele, B. 2019.
\newblock Video object segmentation with language referring expressions.
\newblock In \emph{Computer Vision--ACCV 2018: 14th Asian Conference on Computer Vision, Perth, Australia, December 2--6, 2018, Revised Selected Papers, Part IV 14}, 123--141. Springer.

\bibitem[{Kirillov et~al.(2023)Kirillov, Mintun, Ravi, Mao, Rolland, Gustafson, Xiao, Whitehead, Berg, Lo et~al.}]{kirillov2023segment}
Kirillov, A.; Mintun, E.; Ravi, N.; Mao, H.; Rolland, C.; Gustafson, L.; Xiao, T.; Whitehead, S.; Berg, A.~C.; Lo, W.-Y.; et~al. 2023.
\newblock Segment anything.
\newblock \emph{arXiv preprint arXiv:2304.02643}.

\bibitem[{Li et~al.(2022)Li, Li, Wang, Wang, Qi, Zhang, Liu, Xu, and Lu}]{li2022you}
Li, D.; Li, R.; Wang, L.; Wang, Y.; Qi, J.; Zhang, L.; Liu, T.; Xu, Q.; and Lu, H. 2022.
\newblock You only infer once: Cross-modal meta-transfer for referring video object segmentation.
\newblock In \emph{Proceedings of the AAAI Conference on Artificial Intelligence}, volume~36, 1297--1305.

\bibitem[{Liang et~al.(2021)Liang, Wu, Zhou, Wang, Yang, Wei, and Yang}]{liang2021rethinking}
Liang, C.; Wu, Y.; Zhou, T.; Wang, W.; Yang, Z.; Wei, Y.; and Yang, Y. 2021.
\newblock Rethinking cross-modal interaction from a top-down perspective for referring video object segmentation.
\newblock \emph{arXiv preprint arXiv:2106.01061}.

\bibitem[{Lin et~al.(2017)Lin, Goyal, Girshick, He, and Doll{\'a}r}]{lin2017focal}
Lin, T.-Y.; Goyal, P.; Girshick, R.; He, K.; and Doll{\'a}r, P. 2017.
\newblock Focal loss for dense object detection.
\newblock In \emph{Proceedings of the IEEE international conference on computer vision}, 2980--2988.

\bibitem[{Lin et~al.(2023)Lin, Liu, Zhang, Gao, Qiu, Xiao, Qiu, Lin, Shao, Chen et~al.}]{lin2023sphinx}
Lin, Z.; Liu, C.; Zhang, R.; Gao, P.; Qiu, L.; Xiao, H.; Qiu, H.; Lin, C.; Shao, W.; Chen, K.; et~al. 2023.
\newblock SPHINX: The Joint Mixing of Weights, Tasks, and Visual Embeddings for Multi-modal Large Language Models.
\newblock \emph{arXiv preprint arXiv:2311.07575}.

\bibitem[{Liu et~al.(2019)Liu, Ott, Goyal, Du, Joshi, Chen, Levy, Lewis, Zettlemoyer, and Stoyanov}]{liu2019roberta}
Liu, Y.; Ott, M.; Goyal, N.; Du, J.; Joshi, M.; Chen, D.; Levy, O.; Lewis, M.; Zettlemoyer, L.; and Stoyanov, V. 2019.
\newblock Roberta: A robustly optimized bert pretraining approach.
\newblock \emph{arXiv preprint arXiv:1907.11692}.

\bibitem[{Liu et~al.(2021)Liu, Lin, Cao, Hu, Wei, Zhang, Lin, and Guo}]{liu2021swin}
Liu, Z.; Lin, Y.; Cao, Y.; Hu, H.; Wei, Y.; Zhang, Z.; Lin, S.; and Guo, B. 2021.
\newblock Swin transformer: Hierarchical vision transformer using shifted windows.
\newblock In \emph{Proceedings of the IEEE/CVF international conference on computer vision}, 10012--10022.

\bibitem[{Liu et~al.(2022)Liu, Ning, Cao, Wei, Zhang, Lin, and Hu}]{liu2022video}
Liu, Z.; Ning, J.; Cao, Y.; Wei, Y.; Zhang, Z.; Lin, S.; and Hu, H. 2022.
\newblock Video swin transformer.
\newblock In \emph{Proceedings of the IEEE/CVF conference on computer vision and pattern recognition}, 3202--3211.

\bibitem[{Luo et~al.(2020)Luo, Zhou, Sun, Cao, Wu, Deng, and Ji}]{luo2020multi}
Luo, G.; Zhou, Y.; Sun, X.; Cao, L.; Wu, C.; Deng, C.; and Ji, R. 2020.
\newblock Multi-task collaborative network for joint referring expression comprehension and segmentation.
\newblock In \emph{Proceedings of the IEEE/CVF Conference on computer vision and pattern recognition}, 10034--10043.

\bibitem[{Margffoy-Tuay et~al.(2018)Margffoy-Tuay, P{\'e}rez, Botero, and Arbel{\'a}ez}]{margffoy2018dynamic}
Margffoy-Tuay, E.; P{\'e}rez, J.~C.; Botero, E.; and Arbel{\'a}ez, P. 2018.
\newblock Dynamic multimodal instance segmentation guided by natural language queries.
\newblock In \emph{Proceedings of the European Conference on Computer Vision (ECCV)}, 630--645.

\bibitem[{Milletari, Navab, and Ahmadi(2016)}]{milletari2016v}
Milletari, F.; Navab, N.; and Ahmadi, S.-A. 2016.
\newblock V-net: Fully convolutional neural networks for volumetric medical image segmentation.
\newblock In \emph{2016 fourth international conference on 3D vision (3DV)}, 565--571. Ieee.

\bibitem[{Mo and Tian(2023)}]{mo2023av}
Mo, S.; and Tian, Y. 2023.
\newblock AV-SAM: Segment Anything Model Meets Audio-Visual Localization and Segmentation.
\newblock \emph{arXiv preprint arXiv:2305.01836}.

\bibitem[{Rezatofighi et~al.(2019)Rezatofighi, Tsoi, Gwak, Sadeghian, Reid, and Savarese}]{rezatofighi2019generalized}
Rezatofighi, H.; Tsoi, N.; Gwak, J.; Sadeghian, A.; Reid, I.; and Savarese, S. 2019.
\newblock Generalized intersection over union: A metric and a loss for bounding box regression.
\newblock In \emph{Proceedings of the IEEE/CVF conference on computer vision and pattern recognition}, 658--666.

\bibitem[{Seo, Lee, and Han(2020)}]{seo2020urvos}
Seo, S.; Lee, J.-Y.; and Han, B. 2020.
\newblock Urvos: Unified referring video object segmentation network with a large-scale benchmark.
\newblock In \emph{Computer Vision--ECCV 2020: 16th European Conference, Glasgow, UK, August 23--28, 2020, Proceedings, Part XV 16}, 208--223. Springer.

\bibitem[{Shi et~al.(2018)Shi, Li, Meng, and Wu}]{shi2018key}
Shi, H.; Li, H.; Meng, F.; and Wu, Q. 2018.
\newblock Key-word-aware network for referring expression image segmentation.
\newblock In \emph{Proceedings of the European Conference on Computer Vision (ECCV)}, 38--54.

\bibitem[{Wang et~al.(2023)Wang, Ren, Zhou, Lu, Luo, Shi, Zhang, Song, Zhan, and Li}]{wang2023mathcoder}
Wang, K.; Ren, H.; Zhou, A.; Lu, Z.; Luo, S.; Shi, W.; Zhang, R.; Song, L.; Zhan, M.; and Li, H. 2023.
\newblock MathCoder: Seamless Code Integration in LLMs for Enhanced Mathematical Reasoning.
\newblock \emph{arXiv preprint arXiv:2310.03731}.

\bibitem[{Wang et~al.(2022)Wang, Lu, Li, Tao, Guo, Gong, and Liu}]{wang2022cris}
Wang, Z.; Lu, Y.; Li, Q.; Tao, X.; Guo, Y.; Gong, M.; and Liu, T. 2022.
\newblock Cris: Clip-driven referring image segmentation.
\newblock In \emph{Proceedings of the IEEE/CVF conference on computer vision and pattern recognition}, 11686--11695.

\bibitem[{Wu et~al.(2021)Wu, Jiang, Bai, Zhang, and Bai}]{wu2021seqformer}
Wu, J.; Jiang, Y.; Bai, S.; Zhang, W.; and Bai, X. 2021.
\newblock SeqFormer: Sequential Transformer for Video Instance Segmentation.
\newblock \emph{arXiv preprint arXiv:2112.08275}.

\bibitem[{Wu et~al.(2022)Wu, Jiang, Sun, Yuan, and Luo}]{wu2022language}
Wu, J.; Jiang, Y.; Sun, P.; Yuan, Z.; and Luo, P. 2022.
\newblock Language as queries for referring video object segmentation.
\newblock In \emph{Proceedings of the IEEE/CVF Conference on Computer Vision and Pattern Recognition}, 4974--4984.

\bibitem[{Xu et~al.(2018)Xu, Yang, Fan, Yue, Liang, Yang, and Huang}]{xu2018youtube}
Xu, N.; Yang, L.; Fan, Y.; Yue, D.; Liang, Y.; Yang, J.; and Huang, T. 2018.
\newblock Youtube-vos: A large-scale video object segmentation benchmark.
\newblock \emph{arXiv preprint arXiv:1809.03327}.

\bibitem[{Yan et~al.(2023)Yan, Xu, Zhang, Hong, Chen, Zhang, and Zhang}]{yan2023panovos}
Yan, S.; Xu, X.; Zhang, R.; Hong, L.; Chen, W.; Zhang, W.; and Zhang, W. 2023.
\newblock PanoVOS: Bridging Non-panoramic and Panoramic Views with Transformer for Video Segmentation.
\newblock \emph{arXiv e-prints}, arXiv--2309.

\bibitem[{Yang et~al.(2022)Yang, Wang, Tang, Chen, Zhao, and Torr}]{yang2022lavt}
Yang, Z.; Wang, J.; Tang, Y.; Chen, K.; Zhao, H.; and Torr, P.~H. 2022.
\newblock Lavt: Language-aware vision transformer for referring image segmentation.
\newblock In \emph{Proceedings of the IEEE/CVF Conference on Computer Vision and Pattern Recognition}, 18155--18165.

\bibitem[{Ye et~al.(2019)Ye, Rochan, Liu, and Wang}]{ye2019cross}
Ye, L.; Rochan, M.; Liu, Z.; and Wang, Y. 2019.
\newblock Cross-modal self-attention network for referring image segmentation.
\newblock In \emph{Proceedings of the IEEE/CVF conference on computer vision and pattern recognition}, 10502--10511.

\bibitem[{Zhang et~al.(2021)Zhang, Xie, Barnes, and Li}]{zhang2021learning}
Zhang, J.; Xie, J.; Barnes, N.; and Li, P. 2021.
\newblock Learning generative vision transformer with energy-based latent space for saliency prediction.
\newblock \emph{Advances in Neural Information Processing Systems}, 34: 15448--15463.

\bibitem[{Zhang et~al.(2023{\natexlab{a}})Zhang, Han, Zhou, Hu, Yan, Lu, Li, Gao, and Qiao}]{zhang2023llama}
Zhang, R.; Han, J.; Zhou, A.; Hu, X.; Yan, S.; Lu, P.; Li, H.; Gao, P.; and Qiao, Y. 2023{\natexlab{a}}.
\newblock LLaMA-Adapter: Efficient Fine-tuning of Language Models with Zero-init Attention.
\newblock \emph{arXiv preprint arXiv:2303.16199}.

\bibitem[{Zhang et~al.(2023{\natexlab{b}})Zhang, Hu, Li, Huang, Deng, Li, Qiao, and Gao}]{zhang2023prompt}
Zhang, R.; Hu, X.; Li, B.; Huang, S.; Deng, H.; Li, H.; Qiao, Y.; and Gao, P. 2023{\natexlab{b}}.
\newblock Prompt, Generate, then Cache: Cascade of Foundation Models makes Strong Few-shot Learners.
\newblock \emph{CVPR 2023}.

\bibitem[{Zhang et~al.(2023{\natexlab{c}})Zhang, Jiang, Guo, Yan, Pan, Dong, Gao, and Li}]{zhang2023personalize}
Zhang, R.; Jiang, Z.; Guo, Z.; Yan, S.; Pan, J.; Dong, H.; Gao, P.; and Li, H. 2023{\natexlab{c}}.
\newblock Personalize segment anything model with one shot.
\newblock \emph{arXiv preprint arXiv:2305.03048}.

\bibitem[{Zhang et~al.(2022)Zhang, Zhang, Fang, Gao, Li, Dai, Qiao, and Li}]{zhang2022tip}
Zhang, R.; Zhang, W.; Fang, R.; Gao, P.; Li, K.; Dai, J.; Qiao, Y.; and Li, H. 2022.
\newblock Tip-Adapter: Training-free Adaption of CLIP for Few-shot Classification.
\newblock In \emph{ECCV 2022}. Springer Nature Switzerland.

\bibitem[{Zhou et~al.(2022)Zhou, Wang, Zhang, Sun, Zhang, Birchfield, Guo, Kong, Wang, and Zhong}]{zhou2022audio}
Zhou, J.; Wang, J.; Zhang, J.; Sun, W.; Zhang, J.; Birchfield, S.; Guo, D.; Kong, L.; Wang, M.; and Zhong, Y. 2022.
\newblock Audio--Visual Segmentation.
\newblock In \emph{Computer Vision--ECCV 2022: 17th European Conference, Tel Aviv, Israel, October 23--27, 2022, Proceedings, Part XXXVII}, 386--403. Springer.

\bibitem[{Zhu et~al.(2023)Zhu, Zhang, He, Zhou, Wang, Zhao, and Gao}]{zhu2023not}
Zhu, X.; Zhang, R.; He, B.; Zhou, A.; Wang, D.; Zhao, B.; and Gao, P. 2023.
\newblock Not all features matter: Enhancing few-shot clip with adaptive prior refinement.
\newblock \emph{ICCV 2023}.

\end{thebibliography}

\end{document}